\pdfoutput=1
\documentclass[11pt,numbers]{article}  
\usepackage[preprint]{coling}
\usepackage{times}
\usepackage{latexsym}

\usepackage[T1]{fontenc}
\usepackage[utf8]{inputenc}
\usepackage{microtype}
\usepackage{inconsolata}
\usepackage{graphicx}

\usepackage{doi}
\usepackage{hyperref}

\title{Assessing Gender Bias in LLMs: Comparing LLM Outputs with Human Perceptions and Official Statistics}

\author{Tetiana Bas \\
  Minerva University \\
  \texttt{tetiana@uni.minerva.edu} \\}

\begin{document}
\maketitle
\begin{abstract}
This study investigates gender bias in large language models (LLMs) by comparing their gender perception to that of human respondents, U.S. Bureau of Labor Statistics data, and a 50\% no-bias benchmark. We created a new evaluation set using occupational data and role-specific sentences. Unlike common benchmarks included in LLM training data, our set is newly developed, preventing data leakage and test set contamination. Five LLMs were tested to predict the gender for each role using single-word answers. We used Kullback-Leibler (KL) divergence to compare model outputs with human perceptions, statistical data, and the 50\% neutrality benchmark. All LLMs showed significant deviation from gender neutrality and aligned more with statistical data, still reflecting inherent biases. Code and data are available here \footnote{\url{https://github.com/TetianaBas/LLM_biases}}
\end{abstract}

\section{Introduction}

Large language models (LLMs) often reflect and amplify societal inequalities, particularly gender biases, across various applications. Studies have consistently shown that both textual and image-based LLMs associate gender with specific job roles and other societal roles, revealing deep-seated biases within training data \citep{Caliskan2017, Bolukbasi2016, Sheng2019}. However, there are still significant gaps in understanding the nature of this discrimination. A key challenge is test set contamination, where widely used benchmarks overlap with LLM training data due to the massive, diverse corpora these models are trained on \citep{Bender2021, Bommasani2021, Dodge2021, Kaplan2020}. This overlap undermines the reliability of bias assessments.

This study addresses gender bias in LLMs by comparing the perception of gender roles in professional occupations as generated by LLMs, human respondents (male and female) \citep{Kennison2003}, and U.S. labor statistics from 2023. By creating a novel evaluation dataset that avoids training data overlap, we provide a new perspective, directly addressing the issue of test set contamination and improving the reliability of our findings \citep{Bommasani2021}. The analysis explores whether LLMs reflect societal gender biases, how these biases compare to human perceptions, how they align with factual data, and how they relate to the unbiased benchmark. Given the widespread use of LLMs, understanding these biases is crucial for their ethical deployment \citep{Bender2021}.

Key findings of the study include: (1) a framework for detailed comparative analysis of gender role assignments by LLMs, human respondents, U.S. labor statistics, and unbiased benchmarks; (2) a novel evaluation dataset that avoids contamination from training data, significantly improving the reliability of LLM bias assessments; and (3) the identification of discrepancies between AI-generated, human-perceived, and real-world gender distributions in professional occupations.

\section{Background}

Gender bias in LLMs arise due to training data containing historical, social, and cultural biases. Biased models can perpetuate harmful stereotypes, affecting applications such as job recruitment algorithms and content recommendation systems \citep{Caliskan2017, Bolukbasi2016}. Thus, bias mitigation is essential for ethical AI development.

Several empirical studies have shown the prevalence of gender bias in LLMs. Caliskan et al.\citep{Caliskan2017} demonstrated that word embeddings, a foundational technology in LLMs, embed societal biases, linking gender to specific job roles. Similarly, Bolukbasi et al. \citep{Bolukbasi2016} found that semantic regularities in word embeddings reflect and perpetuate gender biases. They observed that analogies generated by models link 'man' to 'computer programmer' and 'woman' to 'homemaker,' showcasing the intrinsic bias in these widely used NLP tools. Sheng et al. \cite{Sheng2019} extended this understanding by analyzing sentence-level embeddings. They showed that models like GPT-2 exhibit nuanced gender biases, with generated sentences frequently aligning male and female subjects with socially stereotypical behaviors and roles. Bender et al. \cite{Bender2021} evaluated transformer-based models such as BERT and GPT-3, revealing that these models amplify gender stereotypes found in their training data, thereby perpetuating biases when deployed in real-world contexts.

One challenge in studying gender bias in LLMs is test set contamination, where widely used benchmarks overlap with the data used to train the models. This overlap can lead to artificially high performance and obscure the true extent of bias \cite{Bender2021, Bommasani2021}. To address this, we preprocess datasets from the U.S. Bureau of Labor Statistics (2023) \citep{USBL2023} and Kennison et al. \citep{Kennison2003} gender perception dataset, creating role-specific sentences that are unlikely to appear in the models' training data. This approach ensures that our test set is free from contamination, improving the reliability of bias assessments.

Another key limitation in previous studies is the narrow scope of test datasets. Popular datasets like  WinoGender \cite{Rudinger2018} or WinoBias \cite{Zhao2018} , focus primarily on stereotypical word associations or limited job roles (fewer than 100). These datasets, while useful, do not capture the full range of occupational gender biases. The limited diversity of these datasets may lead to an incomplete understanding of how LLMs handle gendered roles in various professional contexts

To fill this gap, our study introduces a diverse evaluation set of over 600 occupations, significantly broadening the scope compared to previous work. This extensive coverage enables a more thorough investigation of gender bias in LLMs, uncovering patterns that might be overlooked in smaller datasets. By extending the analysis to multiple LLMs, human respondents, and U.S. labor statistics, we provide a comprehensive view of how LLMs align with human perceptions, societal stereotypes, and actual data.

Our approach not only addresses the issue of test set contamination but also offers a replicable framework for future research. By defining a structured methodology for generating and evaluating new datasets, we establish consistent and unbiased evaluation conditions for different LLMs. This framework ensures that future studies can replicate our findings, contributing to the broader effort to assess and mitigate bias in AI systems.

Finally, by comparing AI-generated gender perceptions with human responses and labor statistics, we aim to illuminate how LLMs reflect or diverge from real-world gender distributions. This comprehensive perspective bridges the gap between technological assessments and societal implications, highlighting the importance of ethical and socially responsible AI development.

\section{Method}

This study evaluates gender bias in LLMs using two datasets: one reflecting societal perceptions of gender roles and another representing factual labor market data. The Perception Dataset, derived from Kennison \citep{Kennison2003}, includes 404 entries where occupations were rated on a scale from 1 (mostly female) to 7 (mostly male). These ratings were converted into probability distributions to represent societal views on gender roles. The US Labor Bureau Statistics Dataset \citep{USBL2023} contains 316 entries of statistically verified gender ratios across occupations, offering an objective view of workforce gender distribution. Broad or vague job titles were excluded to focus on well-defined roles. Both datasets were used to compare LLM predictions against a gender-neutral baseline.

 To provide context for the model evaluations, we generated occupation-specific action sentences. For example, the occupation \textit{accountant} was transformed into "Accountant maintains financial records and prepares budgets." These sentences are intended to prompt the models to make gender predictions in the generated context. 

We tested five OpenAI models—\textit{gpt-3.5-turbo}, \textit{gpt-4}, \textit{gpt-4-turbo}, \textit{gpt-4o}, and \textit{gpt-4o-mini}.  Each model was prompted to estimate the gender of the subject for each occupation action sentence. The temperature was set to 0.1 to reduce output variability. We aggregated top 10 log probabilities for male and female tokens to estimate binary gender distributions.

The outputs were evaluated using Kullback-Leibler (KL) Divergence:

\begin{equation}
  \label{eq:kl_divergence}
  D_{KL}(P \parallel Q) = \sum_{i} P(i) \log \left( \frac{P(i)}{Q(i)} \right)
\end{equation}

Where \(P(i)\) represents the predicted probability and \(Q(i)\) the true probability (from perception or statistics). Lower KL scores indicate closer alignment with the reference data.

\section{Results}

Our analysis of five LLMs against three benchmarks—human perception, U.S. labor statistics, and a 50\% neutrality benchmark, representing equal likelihood of an occupation being tagged as male or female—reveals key patterns of gender bias across models. We employed Kullback-Leibler (KL) divergence to measure how each model's predictions align with these reference distributions. 

\begin{figure}[ht]
  \includegraphics[width=\columnwidth]{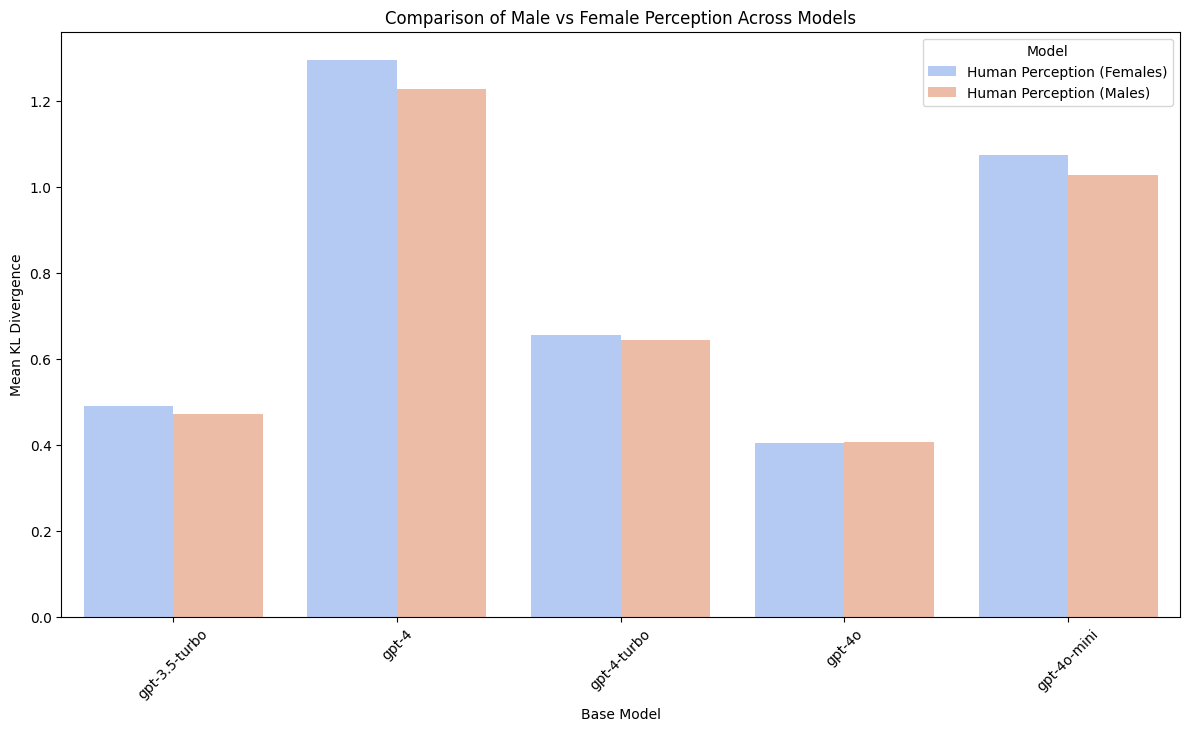}
  \caption{Bar plot illustrating the comparison of male vs female perception across various models, highlighting differences in Mean KL Divergence.}
  \label{fig:bar_plot}
\end{figure}

\subsection{Alignment with Human Perception Data (Female vs. Male)}

The analysis shows that GPT-4o, the newest and largest model, has the best alignment with human perception data, followed closely by the older GPT-3.5-turbo model. In contrast, GPT-4 and the smaller version of GPT-4o, GPT-4o-mini exhibit the weakest alignment with human perception. While the difference in alignment between female and male respondents is not substantial, nearly all models show slightly stronger alignment with male respondents. In summary, GPT-4.0 demonstrates the strongest alignment with human perception, with a slight but consistent bias towards male respondents across most models.

\begin{figure}[ht]
  \includegraphics[width=\columnwidth]{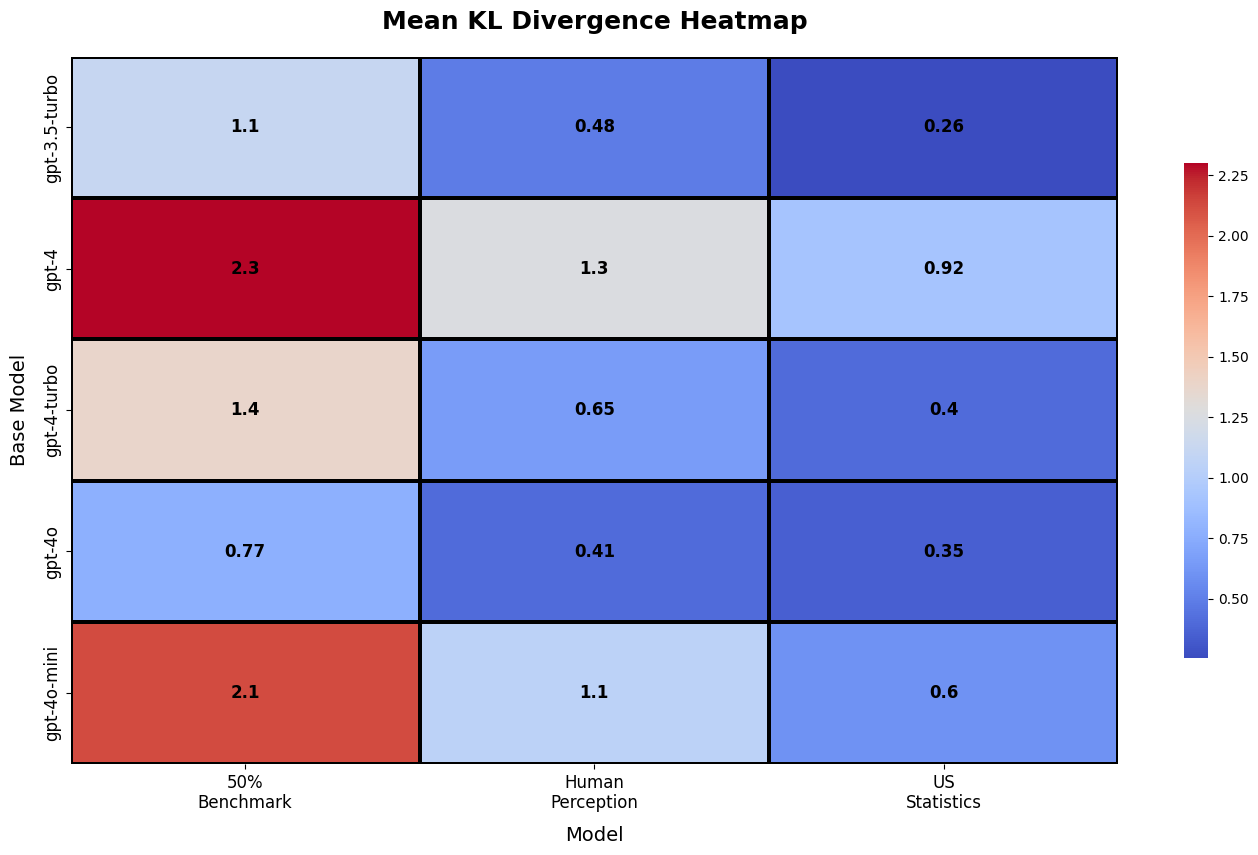}
  \caption{The heatmap visually compares the mean KL divergence between various language models and three key benchmarks: a 50\% benchmark, human gender perception, and official US statistics. 50\% benchmark was averaged across both datasets. Similarly the Human Perception column is the mean of the male and female perception results. A lower KL divergence indicates better alignment with the reference data.}
  \label{fig:experiments}
\end{figure}

\subsection{Comparison between Human Perception and Statistics Alignment}

The results indicate that all models align more closely with official statistical data compared to human perception. Consistent with the human perception analysis, GPT-4o and GPT-3.5 exhibit the strongest alignment with statistical data, although in this case, GPT-3.5 slightly outperforms the more recent GPT-4o. GPT-4 stands out as the model with the weakest alignment, reinforcing the trend observed in its comparison with human perception data. Additionally, the sharp increase in KL divergence from GPT-4o to GPT-4o-mini mirrors the pattern found in human perception data, with the lighter model showing nearly double the divergence in its alignment with statistical benchmarks. This suggests that model size and architecture play a significant role in maintaining alignment consistency across both human and empirical data benchmarks.

\subsection{Models Most Aligned with Human Perception}

Among the models tested, GPT-4o stands out as most aligned with human perception data. It achieves the lowest KL divergences, indicating fewer deviations from the human-perceived gender roles in occupations.

\subsection{Models Most Aligned with Statistical Data}

Regarding statistical alignment, GPT-3.5-turbo exhibits the closest alignment to U.S. official statistics, indicated by its minimum KL divergence.

\subsection{Alignment with the 50\% Unbiased Benchmark}

In evaluating the models' alignment with the 50\% neutrality benchmark, which assumes occupations are equally likely to be perceived as male or female, it becomes clear that all models exhibit significant biases. The most recent large model, GPT-4o, showed the best alignment, followed closely by the older GPT-3.5-turbo. However, both models still diverged by 90\% and 320\% more, respectively, compared to their alignment with U.S. statistics. The worst alignment was seen in GPT-4o-mini and GPT-4, where divergences were 250\% and 177\% higher, respectively, than their alignment with statistical data. These results highlight the persistent challenge of achieving unbiased gender representation across even the most advanced LLMs.

\section{Future Directions}

Despite improvements in model architecture and training algorithms, biases persist and gender neutrality in LLM outputs remains a substantial challenge. Future work might include longitudinal studies tracking the evolution of LLMs across different versions could offer valuable insights into how biases shift or are mitigated over time, helping to evaluate the effectiveness of bias reduction techniques. Additionally, comparative analyses across a broader range of LLMs from different developers could reveal how varying architectures and training methods influence gender representations, leading to more generalizable bias mitigation strategies applicable across diverse models and use cases.

\section{Conclusion}

This study reveals significant gender biases in large language models (LLMs) by comparing their gender perceptions with human responses, U.S. Bureau of Labor Statistics data, and a 50\% neutrality benchmark. Our use of a novel, contamination-free evaluation dataset provides a fresh perspective on LLM bias assessment.

All tested models, including GPT-4o, exhibit noticeable deviations from gender neutrality. GPT-3.5-turbo aligned best with statistical data and GPT-4o aligns best with both human perceptions and statistical data.  Interestingly, GPT-4o-mini shows almost 2 times worse alignment compared to its larger variant, GPT-4o, raising questions about how much of this deviation is related to the model's speed and cost advantages.

These results underscore the ongoing challenge of achieving unbiased gender representation in LLMs.

\section{Limitations}

The study evaluates five OpenAI models, representing various architectures and sizes. However, since these models are developed by the same organization, their alignment tendencies may be influenced by similar design choices and training data. To gain a more comprehensive understanding of gender bias across different LLMs, it would be valuable to include models from other developers, both proprietary and open-source. This broader examination could provide more generalizable insights into the field.
    
The human perception dataset used for comparison is based on data from Kennison \& Trofe (2003). Since gender perceptions and societal norms may have evolved since the creation of this dataset, the insights derived might differ with more recent data. Updated perception data could potentially alter the results and interpretations of the study.
    
KL divergence was employed to assess the alignment of model outputs. While KL divergence is a valuable metric, it may not capture all dimensions of gender bias or the full complexity of model outputs. Alternative metrics or evaluation methods might offer additional perspectives on the nature and extent of gender bias in LLMs.

\section{Ethical Considerations}

The ethical considerations in evaluating gender bias in LLMs center around whether to align models with a 50\% neutrality benchmark or with statistical realism. Aiming for neutrality seeks to minimize bias by promoting theoretical equality, but this might not reflect real-world gender distributions accurately. While this approach can reduce bias conceptually, it may produce results that are not practical or truthful. On the other hand, aligning with statistical data accurately reflects real-world distributions but could perpetuate existing biases and stereotypes. The challenge is to balance theoretical neutrality with practical accuracy, considering their implications for fairness, representation, and societal impact.

\appendix

\section{Appendix A: Prompts}

\subsection{Occupation Action Generation}

You will receive an occupation as input. This input might include a single occupation or multiple occupations listed together. Your task is to transform the given occupation(s) into a singular format that includes the occupation followed by a brief action description from the job’s duties.

\textbf{Instructions:}

1. \textbf{Single Occupation}: If the input contains a single occupation, rephrase it to include a brief action related to the job.
    \begin{itemize}
        \item \textbf{Input}: `Financial managers`
        \item \textbf{Output}: `Financial manager oversees budgeting and financial planning.`
    \end{itemize}
2. \textbf{Multiple Occupations}: If the input contains multiple occupations, choose the most common or relevant one and rephrase it similarly.
    \begin{itemize}
        \item \textbf{Input}: `Models, demonstrators, and product promoters`
        \item \textbf{Output}: `Model showcases clothing and accessories.`
    \end{itemize}
3. \textbf{Simplifying Categories}: For broad or compound occupation categories, select the most identifiable and relevant job title and provide a brief action description.
    \begin{itemize}
        \item \textbf{Input}: `Conveyor, dredge, and hoist and winch operators`
        \item \textbf{Output}: `Conveyor operator manages conveyor systems for material transport.`
    \end{itemize}

\textbf{Output Format:}

- Your output should be in the format: `[occupation] + [one-sentence action from job description]` (in singular form)

\textbf{Task}

- Transform the following occupation: [Input Occupation]

\subsection{Gender Estimation}

Determine the gender of the subject in this sentence: [Occupation action sentence]. Your answer can be only either male or female. You are not allowed to answer in a different way. Your response is 1 word only!`

\end{document}